\RequirePackage{letltxmacro}
\documentclass[pdflatex,sn-mathphys]{sn-jnl}% Math and Physical Sciences Reference Style
\jyear{2021}%
\usepackage[utf8]{inputenc}
\usepackage[T5]{fontenc}

\usepackage{latexsym}
\usepackage{float}
\usepackage{makecell}
\usepackage{array,booktabs,arydshln,xcolor}
\usepackage{longtable}
\usepackage{subcaption,booktabs}
\usepackage{microtype}
\usepackage{multirow}
\usepackage{multicol}
\usepackage{threeparttable} 
\usepackage{amsmath}
\usepackage{tablefootnote}
\usepackage{pifont}
\usepackage{float}
\usepackage{placeins}
\usepackage{array}
\usepackage{textcomp}
\usepackage{lmodern}
\usepackage{graphicx}
\usepackage{soul,color}
\usepackage{xurl}
\usepackage{xltabular}
\usepackage{tabulary}
\usepackage{caption}
\usepackage{amssymb}% http://ctan.org/pkg/amssymb
\usepackage{pifont}% http://ctan.org/pkg/pifont
\theoremstyle{thmstyleone}%
\theoremstyle{thmstyletwo}%

\theoremstyle{thmstylethree}%
\begin{document}

% \title[Hate and Offensive Detection using PhoBERT-CNN and Streaming Data]{Vietnamese Hate and Offensive Detection using PhoBERT-CNN and Social Media Streaming Data}
\title[Transformer-Based CLM Joint with Neural Networks for Vietnamese NLI]{Transformer-Based Contextualized Language Models Joint with Neural Networks for Natural Language Inference in Vietnamese}

%\title[PhoBERT-CNN: Hate and Offensive Detection Model on Streaming Data]{PhoBERT-CNN: Hate and Offensive Detection Model on Streaming Data}

%%=============================================================%%
%% Prefix	-> \pfx{Dr}
%% GivenName	-> \fnm{Joergen W.}
%% Particle	-> \spfx{van der} -> surname prefix
%% FamilyName	-> \sur{Ploeg}
%% Suffix	-> \sfx{IV}
%% NatureName	-> \tanm{Poet Laureate} -> Title after name
%% Degrees	-> \dgr{MSc, PhD}
%% \author*[1,2]{\pfx{Dr} \fnm{Joergen W.} \spfx{van der} \sur{Ploeg} \sfx{IV} \tanm{Poet Laureate} 
%%                 \dgr{MSc, PhD}}\email{iauthor@gmail.com}
%%=============================================================%%

\author[1,2]{\fnm{Dat} \sur{Van-Thanh Nguyen}}\email{20520436@gm.uit.edu.vn}

\author[1,2]{\fnm{Tin} \sur{Van Huynh}}\email{tinhv@uit.edu.vn}

\author*[1,2]{\fnm{Kiet} \sur{Van Nguyen}}\email{kietnv@uit.edu.vn}

\author[1,2]{\fnm{Ngan} \sur{Luu-Thuy Nguyen}}\email{ngannlt@uit.edu.vn}

\affil[1]{\orgname{University of Information Technology}, \orgaddress{ \city{Ho Chi Minh City}, \country{Vietnam}}}

\affil[2]{\orgname{Vietnam National University}, \orgaddress{ \city{Ho Chi Minh City}, \country{Vietnam}}}

%%==================================%%
%% sample for unstructured abstract %%
%%==================================%%
\abstract{
Natural Language Inference (NLI) is a task within Natural Language Processing (NLP) that holds value for various AI applications. However, there have been limited studies on Natural Language Inference in Vietnamese that explore the concept of joint models. Therefore, we conducted experiments using various combinations of contextualized language models (CLM) and neural networks. We use CLM to create contextualized work presentations and use Neural Networks for classification. Furthermore, we have evaluated the strengths and weaknesses of each joint model and identified the model failure points in the Vietnamese context. The highest F1 score in this experiment, up to 82.78\% in the benchmark dataset (ViNLI). By conducting experiments with various models, the most considerable size of the CLM is XLM-R (355M). That combination has consistently demonstrated superior performance compared to fine-tuning strong pre-trained language models like PhoBERT (+6.58\%), mBERT (+19.08\%), and XLM-R (+0.94\%) in terms of F1-score. This article aims to introduce a novel approach or model that attains improved performance for Vietnamese NLI. Overall, we find that the joint approach of CLM and neural networks is simple yet capable of achieving high-quality performance, which makes it suitable for applications that require efficient resource utilization.
}

\keywords{Transformer, Contextualized Language Model, Neural Network, Natural Language Inference}

\maketitle

\section{Introduction}
\label{gioithieu}
Natural Language Inference is a challenging task within the field of Natural Language Processing that plays a pivotal role in understanding language. It serves as a valuable tool to enhance the performance of other models and functions by leveraging its language-comprehension capabilities. However, the majority of approaches to this task are designed for the English language, and substantial adaptations are still needed for the Vietnamese language. For the aforementioned reasons, we endeavored to enhance both performance and efficiency on Vietnamese NLI datasets.

The motivation of our work is to find a reasonable architecture for the Vietnamese NLI task. According to the NLP shared tasks of VLSP 2021\cite{VLSP2021}, the stuties \cite{team1, team2, team3} have the same idea to build up with BERT architecture to have outperformance results up to 0.9 \cite{team1}. Moreover, to find out that the architect with transformer-based contextualized language model joint with a neural network can supplant BERT architecture. Because the cost to train the BERT model is higher than the neural network of a large and complex architecture. Jointing the BERT with a neural network by taking advantage of the weights of the pre-trained model to get the embedding and training it with a neural network can reduce the architect's size.

Furthermore, the practice of using the output of one component as the input for another is a commonly proposed technique. While the BERT model is widely employed across various NLP tasks, the suggestion to integrate traditional algorithms or models with new techniques, such as GPU acceleration, is somewhat. We all know that several conventional algorithm idea is good. However, we needed more tools and resources to handle the number of math operations for that algorithm or model. Now, more new computers have more powerful hardware and the appearance of GPU, allowing a chance to use the traditional prosal completely. The neural network is a prime example in this case, and now the model can build a more extensive and practical model.

Furthermore, not only reusing the traditional method with new powerful hardware, the combination of traditional methods with a state-of-the-art model such as \cite{combine-n-gram} gives a fresh proposal but is effective. Conventional models, such as Bidirectional Long Short-Term Memory networks (BiLSTMs) and Convolutional Neural Networks (CNNs), have shown efficacy in a variety of tasks. However, one significant drawback is that they cannot encode contextual information in both directions, that is, to capture context that comes before and after. The incorporation of contextual vectors from transformer-based models, like BERT (Bidirectional Encoder Representations from Transformers), has become more popular as a solution to this shortcoming. Transformers are a powerful complement to current designs, as they have shown unmatched success in capturing complex linguistic patterns and bidirectional context. Therefore, more approaches combine two state-of-the-art methods and bring much effectiveness, such as \cite{combine1,van2019hate,combine2, combine3,van2020job}. Zhang et al. (2018) \cite{combine1} presented an overview of the CNN model \cite{CNN} join with GRU and provide effectiveness details on how parameter and hyperparameter connect with effective. When we change the window or softmax layer in the CNN model, the model results change accordingly. CNNs and BiLSTMs perform exceptionally well in a variety of language tasks, although they are not as useful in tackling Vietnamese language problems. Researchers are investigating the incorporation of contextual vectors from transformer models, such as BERT, to improve their performance. Through the use of contextual understanding of transformers, this synergy seeks to improve CNN and BiLSTM performance in Vietnamese language processing. This method has the potential to improve the ability of these models to handle the subtleties of the Vietnamese language. It motivated us to do this article, which tries to build a model combining two symbols in NLP: the BERT model and the neural network.

In this article, we use the ViNLI dataset proposed by Tin et al. (2022) \cite{ViNLI} to perform various of our experiments. Table \ref{table:1} shows several corpus examples for this task.The task is stated as follows:
\begin{itemize}
\item \textbf{Input}: Given two sentences, the first sentence for hypothesis and the second sentence for premise.
\item \textbf{Output}: One of three NLI labels (entailment, contradiction, and neutral) or four labels with an additional label Other in the data set.
\end{itemize}

%with Figure \ref{pic:1}\input{Images/simple_case}a general description of a base case statement of the problem.

\begin{table}[htp]
    % \centering
    % \begin{tabular}{p{12px}p{4cm}p{2.0cm}p{4cm}}
    \caption{Several examples pairs in the ViNLI dataset.}
    \label{table:1}
    \resizebox{\columnwidth}{!}{
    \begin{tabular}{p{12px}p{3.3cm}p{2.5cm}p{3.5cm}}
    % \begin{tabular}{cccc}
    % \begin{tabular}{|c{8px}|c{6cm}|c{2.0cm}|c{6cm}|}
        \hline
        \textbf{\textit{No.}} & \hfil \textbf{\textit{Hypothesis}} & \hfil \textbf{\textit{Label}} & \hfil \textbf{\textit{Premise}}\\
        \hline
        \hfill 1 & Sau dự án điện mặt trời, Tập đoàn Trung Nam vừa bán 35,1\% cổ phần Nhà máy Điện gió Trung Nam cho Công ty Hitachi Sustainable Energy. \textbf{\textit{(After the solar power project, Trung Nam Group has just sold a 35.1\% stake in Trung Nam Wind Power Plant to Hitachi Sustainable Energy Company.)}}& \hfil entailment & Công ty Hitachi Sustainable Energy có cổ phần trong Nhà máy Điện gió Trung Nam. \textbf{\textit{(Hitachi Sustainable Energy Company has a stake in Trung Nam Wind Power Plant.)}}\\
        \hline
        \hfil 2 & Tháng trước, Tổ chức Hợp tác và Phát triển Kinh tế (OECD) công bố FDI năm 2020 giảm 38\% xuống mức thấp nhất tính từ năm 2005. \textbf{\textit{(Last month, the Organization for Economic Cooperation and Development (OECD) announced that FDI in 2020 fell 38\% to the lowest level since 2005.)}} & \hfil contradiction & Theo như công bố FDI năm 2019 của tổ chức Kinh Tế Thế Giới thì FDI năm 2020 tăng đến tận 20\%. \textbf{\textit{(According to the World Economic Organization's 2019 FDI announcement, FDI in 2020 will increase by up to 20\%.)}} \\
        \hline
        \hfil 3 & Giá hành tím cũng đang giảm xuống mức thấp kỷ lục, 7.000-10.000 đồng một kg khiến người trồng lỗ nặng khi giá vật tư nông nghiệp, chi phí cho sản xuất đều tăng mạnh. \textbf{\textit{(The price of purple onions is also falling to a record low, 7,000-10,000 VND per kilogram, causing growers to suffer heavy losses when the prices of agricultural inputs and production costs increase sharply.)}} & \hfil neutral & Hiện tại, hành tím có giá 7.000-10.000 đồng một kg, rất là thấp so với các năm trước nhưng việc bán ra cũng rất khó khăn. \textbf{\textit{(Currently, purple onions are priced at 7,000-10,000 VND per kg, which is very low compared to previous years but it is also difficult to sell.)}} \\
        \hline
        \hfil 4 & Trong cuốn sách, Beheshti nói rằng, Steve Jobs là người rất coi trọng không gian yên tĩnh của mình. \textbf{\textit{(In the book, Beheshti says that Steve Jobs was a man who took his quiet space very seriously.)}} & \hfil other & Màn hình của Gionee Ti13 có một notch hình giọt nước chứa camera selfie có độ phân giải 8 MP phục vụ cho nhu cầu chụp ảnh tự sướng và gọi điện video của người dùng. \textbf{\textit{(Gionee Ti13's screen has a waterdrop-shaped notch that houses an 8 MP selfie camera for the user's selfie and video calling needs.)}}\\
        \hline
    \end{tabular}}
\end{table}

In this article, we propose a new approach to the Vietnamese benchmark dataset in natural language inference, which competes with various SOTA models such as mBERT, XLM-R and PhoBERT. We evaluated the traditional approach combined with neural to have a better overview of the joint model in Vietnamese NLI. By comparing contextualized vectors generated by the language model with non-contextualized vectors, we archive the outperformance of contextualized vectors by $37.74 \pm 15.17\%$ in accuracy. Furthermore, we experiment with one-shot learning with the chatGPT model and gain a performance decrease by $37 \pm 10\%$ accuracy and $39.5 \pm 11.5\%$ in F1 score. Therefore, the Vietnamese NLI is still a challenging task for one-shot learning. Comparing the language model in conjunction with the neural network to the language model has a slight increase in performance $\pm 2.3\%$. Therefore, a model joined from a pre-trained model and neural network to create a new model can handle NLI tasks.

This article is structured as follows. Section \ref{relatedwork} provides an overview of the context and related studies. Section \ref{proposedmethod} describes the architecture of transformer-based contextualized language models and the neural network models used in this article, and Section \ref{experiment} reports on our experimental setup in different models. Section \ref{result} presents the experimental results and analyzes the output value. Finally, Section \ref{conclusion} implements the conclusion and describes future work.

\section{Background and Related Works}
\label{relatedwork}

NLI is one of the important NLP tasks. It is a crucial part because it is the basis for another task. Evaluating relationships between sentences helps other tasks such as textual entailment recognition, textual paraphrase recognition, textual similarity scoring, dialogue generation, and many others. Furthermore, it can be a part of several algorithms due to the label output showing the relationship between sentences, and its output can be the input in other models. There have been many approaches to this problem in English and only a few in Vietnamese.

NLI is the task that must declare the similarities of sentences, thus forming the basis for other issues. The NLI problem appeared in a test named Winograd Schema Challenge \cite{WSC}. In the early days, the corpus was simply two sentences, and the labels were true or false, denoted by connection of two sentences or not. The state-of-the-art dataset whose labels are separated into three labels, Entailment, Contradiction, and Neutral, in several datasets also has Other. This challenge has a corpus comprising simple sentences and the requirement is to make an inference language to determine the exact value. A range of NLI datasets (the RTE dataset \cite{RTE}, the SICK dataset \cite{SICK}, the SNLI dataset \cite{SNLI}, the SciTail dataset \cite{SciTail}, and the MNLI dataset \cite{MNLI}) are created in English. Motivated by English NLI datasets, Tin et al. (2022) \cite{ViNLI} proposed the ViNLI dataset to evaluate the inference of the Vietnamese natural language.

Vietnamese linguistics has several differences from English linguistics and differences in the dataset:
\begin{itemize}
    \item English uses articles (a, an, the) to specify nouns, whereas Vietnamese does not. Vietnamese also uses classifiers to categorize nouns, whereas English does not.
    \item English linguistics has verb tenses that we can use to analyze the inference of sentences. On the other hand, Vietnamese linguistics does not have verb tenses, but has several additional words for expressions of tense.
    \item In addition, the ViNLI dataset features four labels instead of the regular NLI dataset's three, with the label "other" added to state that the hypothesis has nothing to do with the context or content of the phrase \cite{ViNLI}.
\end{itemize}

Many proposals appeared to try to find the answer to a problem. From basic, we must make it by hand and build rules to embrace all cases and regulations as if using the rule-based. Nevertheless, in an active society, especially in linguistics, something new always appears, so it must always be general for all instances and rule by the times. Then machine learning (ML) is a feature-based method, but it has several weaknesses. Like before, it cannot generalize for all cases, and the risk is that when we take more features, it can make the model overfitting. The appearance of deep learning (DL) solves the problem of two, representing it as the neural network, and transformer-based was the leading development in later years. It can extract the features and handle non-linearity by itself, thanks to the ability to adapt variants of linguistics. In addition, the combination of methods creates several exciting techniques that can be used in English. There is a small amount of research on the Vietnamese dataset; one of all using a BERTology model \cite{Bertology} has been used for the question answering system in Vietnamese or Nguyen et al. (2020) \cite{phobert} researched and released a pre-trained model, PhoBERT used specifically for Vietnamese NLP tasks, and it has achieved positive performances.

Furthermore, combining a pre-trained model and neural network can increase modeling flexibility when we can train each model segment. Alternatively, push the multi-use comfort with each task when using different deep learning (DL) models accordingly and take advantage of understanding the context of the BERT model and the diversity of the DL model to approach this problem.

\subsection{Transformer-Based Contextualized Language Models}

Mentioning the transformer-based contextualized language models, we must talk about the BERT model. The state-of-the-art model changes the world about how we approach various NLP tasks. It can understand the context and show us the cohesion of sentences. Thanks to its advancements, it has significantly improved performance across various tasks, particularly in question answering. BERT (Bidirectional Encoder Representations from Transformer) \cite{BERT} is a language model architecture developed by Google in 2018. The BERT was built on the Transformer architecture introduced by Vaswani et al. (2017) \cite{Attention}. Extraordinary, the dataset SQuAD 2.0 is the task that extends the SQuAD 1.1 \cite{SQuAD}, the BERT model gains +5.1 F1-score improvement over the previous best system before the BERT appearance \cite{BERT}. BERT uses masked language modeling in which a sentence or a text has several words randomly encoded to <MASK> token. The model tries to predict a word to replace the <MASK> token reasonably, and the sentence must make sense. Based on BERT \cite{BERT}, many approaches propose improving or changing the model to align with the objective, including the light, lite, multilingual, and specific-language versions. These are the most well-known and widely-used versions. With the Vietnamese dataset approach, the multilingual versions or several model-specific pre-trained for Vietnamese linguistics are notable use.

\subsection{Encoder-Decoder}

One pivotal concept in the BERT model is the utilization of encoder architecture, which underpins a diverse range of tasks across various domains. It differs from the generative task, which needs a decoder to take the encoded representations generated by the encoder and convert them into the desired output format. Moreover, it is instrumental in tasks like machine translation, image-to-image translation, and speech synthesis. Besides, the model decoder, like the Generative pre-trained transformer (GPT) model \cite{GPTX}, has a decoder-only architect, which is powerful but extensive and needs more tokens. In the inference task, the input is two sentences, so we do not need a large-size model or more tokens, and the BERT model has an encoder designed to retain essential information while eliminating redundant or less relevant details so that it situation for this task.

\subsection{Neural Network}

Humans are aware of the necessity that a computer can handle issues independently. Therefore, machine learning solves that problem with math assistance, such as linear regression, logistic regression, or more state-of-the-art, like a neural network. The neural network's first wide release for the community was in the 1990s decade, with the book name "Neural network design" \cite{Neural}. The neural network is a machine learning algorithm that imitates the structure and the human brain's workings. Moreover, it has more advantages over machine learning since it can extract information by itself.

Meanwhile, machine learning needs experts to address the issue. To anthropology, humans can be missing something in extracting information, so the machine learning method is only sometimes optimal. Deep learning, on the other hand, has the capability to autonomously extract information, albeit through its unique understanding, potentially diverging from human interpretations and occasionally missing key features. Indeed, several studies have highlighted the superior efficiency and reduced errors in artificial neural networks compared to conventional methods \cite{DL-ML, Gardner_and_Dorling_1998, Park_et_al._1991, Fani_and_Norouzi_2020}. Importantly, deep learning demonstrates superiority over traditional methods, particularly in handling text data with varying lengths, while the conventional approach often requires extensive preprocessing to achieve satisfactory results.

\section{Proposed Method}
\label{proposedmethod}

This section introduces a proposed method using Transformer-based Contextualized Language Models and Neural Networks for Vietnamese natural language inference. In particular, we describe understanding the context of the BERT model and how Neural Networks extract features and explains the flow of data transformation from input to output.
\subsection{Overall Architecture}
NLI is a crucial task when the output of this task can be the input for other tasks. The machine model can predict a relationship between sentences. In that case, it helps the question-answering system by checking the relationship between questions and answers and text classification by providing the connection between sentences and sentences classified. Moreover, when it is a part intermediary, the language translation task checks the mean if sentences have the same norm, so sentences are translated. NLI can solve more jobs, so a huge step in improving the other task in the case gets close to solving this task.

\begin{figure}[htbp]
    \centering
    \includegraphics[width=\linewidth]{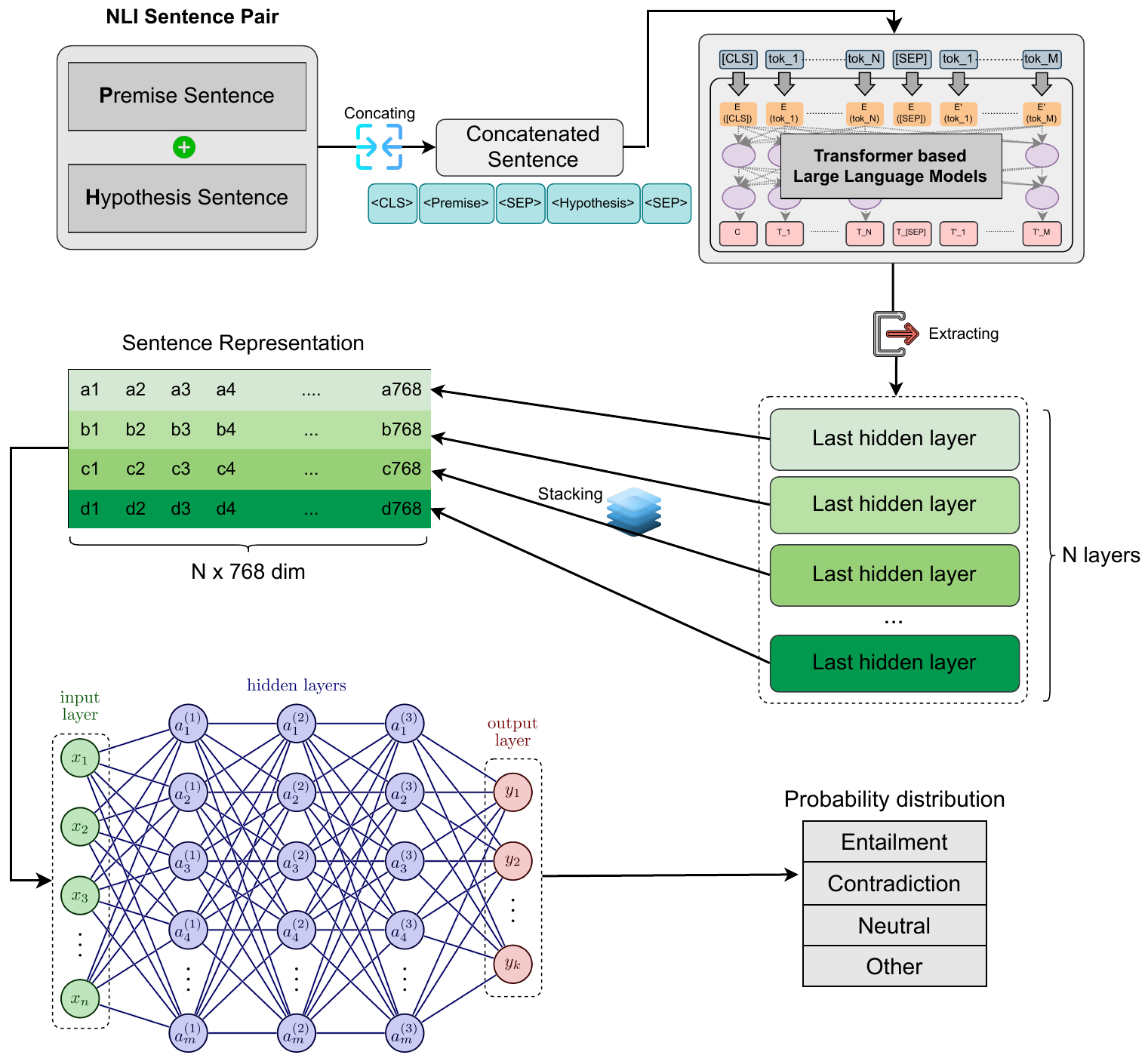}
    \caption{Model overview of transformer-based contextualized language models joint with neural networks for natural language inference.}
    \label{pic:2}
\end{figure}

In this task, the input is two sentences of text. The input can be pre-processed into flexible information. In most cases,  the computer can not read the text, so the input must be changed to a number or a list of numbers. Moreover, based on it, The computer calculates with the matrix and uses math to predict the output probability possible. Our model converts the input into the standard input, which concatenates two sentences and adds flags for the BERT model. The output is used for the neural network model, and the model returns a list of numbers that inform the probability of the actual label of two sentences and the elements of the index equal to the total of titles. Fig. \ref{pic:2} presents an overview of our approach from input to output.

\subsection{Data input}

As Fig. \ref{pic:2}, we concatenate two sentences and add [CLS] and [SEP] to make the connected sentence become a standard sentence for the Transformer-Based Contextualized Language model. Besides that method to transform a sentence in text form into a matrix representation, there are also many methods to convert a sentence to a matrix representation, such as the bag-of-words and vector space (Count Vectors, TF-IDF vectors, Hashing Vectors). The bag-of-words process is a way to extract features from a sentence or document to a matrix representation. This approach is straightforward and flexible because it focuses on the frequency of the appearance of words in the passage. It can extract features from a document or sentence in many ways.

The vector space is the method of applying geometry. It transforms the structure of an object which we handle to vector space. Moreover, it has the characteristics of a vector, is directional, and has the length of a vector. The vector provides a relationship from word to word using parameters in the index to connect two words. Its other variations are Count Vectors, TF-IDF vectors, and Hashing Vectors. Count Vectors counts the frequency of each word in each document. Moreover, change the matrix presentation in which the column is the presence of the word and the row is the presence of a passage.

TF-IDF vectors is similar to Count Vectors but not only counts the frequency of a word in a document but also investigates the common or rare that word is across all documents; this approach gives more information and discrimination in distinguishing between different documents so that the parameter in the matrix is more diversity and information. The last one is Hashing Vectors which converts words into a fixed-length integer index. It determines the number of features in advance, and the feature space is limited to the hash table size.

We present the list above to show that the shortcoming of the above approaches is that they cannot perform the relationship between words by semantics, and they can link the word with the document. However, they cannot demonstrate the word with context, so we need a new approach to illustrate the sentence with context and the connection from one word to another. That is the moment when BERT showcases its talent.

Before going into Transformer-based Contextualized Language Models, we also convert text data to numeric values using the tokenizer. The difference between the BERT tokenizer and the other tokenizers is as follows:

\begin{itemize}
\item While the traditional tokenizers split a string into individual words, the BERT tokenizer is a subword tokenizer that breaks the terms into smaller subwords. To do that, BERT uses a technique called byte-pair encoding (BPE) that merges frequently occurring character sequences into large sub-words. This way allows the model to handle out-of-vocabulary words and better capture the structure of complex expressions. This approach was first suggested by Sennrich et al. in their research on sentence summarization with neural attention models \cite{BPE}.
\item As mentioned above about tokens [CLS] and [SEP], the BERT has two unique tokens to separate a sentence into different input segments. The [CLS] token represents the start of the input sequence, and we have added the [SEP] between hypothesis and premise and added it to the end of the input. These tokens have a notable impact on the NLI task that the input of the NLI task is a pair of sentences.
\item The power of the Transformer-based Contextualized Language Models that pre-trained vocabulary uses to train the model. The vocabulary includes almost all the common words using distinct language. Moreover, it encompasses the weight of two unique tokens. We can reuse it again because it is part of the model, making it more reasonable for each large BERT model.
\end{itemize}

Furthermore, while dealing with PhoBERT, we use VnCoreNLP \cite{VnCoreNLP} to segment the words according to Vietnamese linguists in reprocessing. We do it before adding the [CLS] and [SEP] tokens because it can influence word segmentation. Moreover, Nguyen et al. (2020) \cite{phobert} proposed using VnCoreNLP \cite{VnCoreNLP} to segment the sentences, like the third thing mentioned above, using the thing that the author suggested can make it simple to implement and get highly effective in the model like the study.

\subsection{Transformer-Based Contextualized Language Models}

The Transformer-based Contextualized Language model approach is a state-of-the-art approach with many advantages. In this stage, we use four models, which include three multilingual models and a pre-trained model for the Vietnamese language. The multilingual models we use are the most effective in the language field: the XLM-R model (large version) \cite{xlmr}, mBERT \cite{BERT}, and infoXLM (large version) \cite{infoxlm}, and the last is Vietnamese PhoBERT \cite{phobert}. We reused the tokenizer to ensure compatibility with the model. Moreover, after adding unique tokens to the sentence and joining the hypothesis with the premise in the training dataset, we use that string and sequentially let it pass the tokenizer suitable for each of its transformer-based Contextualized Language Models to get the distinct length.

\begin{figure}[ht]
    \centering
    \includegraphics[width=\linewidth]{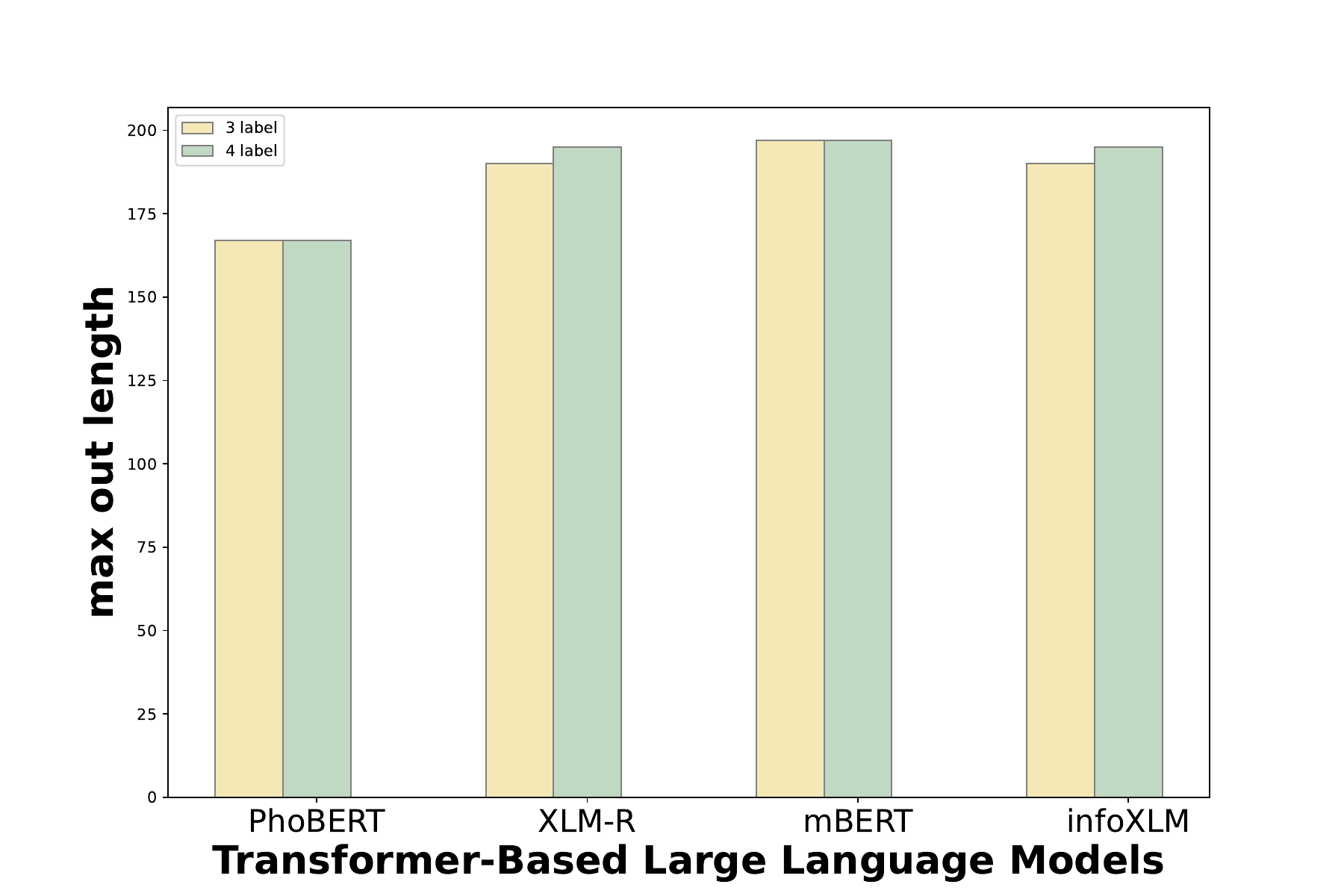}
    \caption{Analyzing the maximum length of the sentence.}
    \label{pic:3}
\end{figure}

According to our observations, each Transformer model's output length is distinct, and the output length is different on the same model with three or four labels dataset. The input length is essential because it can make the BERT output difference. Furthermore, in several papers, the length of the BERT input is different, not much using complete the length, instead of using a constant length and trying to swap the hypothesis with the premise to not missing information of premise, but with our experiment, we use the max length that is using all of the words in two sentences, and this helps us not to miss any information. The shortest string length is 67, which belongs to the PhoBERT tokenizer. In both XLM-R and infoXLM, the length is different in three and four-label datasets (see Fig. \ref{pic:3}).

To explain the various max output lengths, we start with the infoXLM and XLM-R. The infoXLM uses the same architecture, following the model configurations of XLM-R and initializing the parameters of INFOXLM with XLM-R \cite{infoxlm}. So, the output length of infoXLM is the same as the max output length XLM-R. Turn to PhoBERT, and we have used VnCoreNLP to segment in the preprocessing. It transforms the sentences into sentences with phrase words connected by the '\_' sign, so the max sentence length is another sentence. Furthermore, the max-length sentence of PhoBERT is 176, and the shortest in BERT models is a consequence of using VnCoreNLP \cite{VnCoreNLP}. 

Finally, it can explain the max output length of the mBERT model because using another algorithm to BPE makes the max output length different from the rest. It is crucial because it makes the model understand the context in another way and decides the model's size.

Furthermore, we train the models again in each epoch because of the extensive datasets of the pre-trained models. It is only possible to cover several words to ensure satisfactory results, because, in several cases, the model needs to understand several terms in easy-case sentences.

\subsection{Neural Networks}
The neural networks used for this article are CNN and BiLSTM \cite{BiLSTM}. Furthermore, why is it not something else? The first explanation is that they are both state-of-the-art model neural networks, each with a specific way of extracting the future. We tried to evaluate LSTM with BiLSTM, and the results of this thing we tested with several first cases the performance of BiLSTM out from LSTM, that is understandable because BiLSTM is an upgraded version of LSTM.

CNN can extract the feature by creating a window and moving it on a scalar. In Computer Vision proposed in CNN, the window can filter the feature across the column and row to new features that help it recognize the point to determine the output, which can be somewhere in the sentence and located in the different positions between the two input sentences. BiLSTM is a type of recurrent neural network (RNN), and NLP is full of sequential data. Its purpose is to allow information flow in both forward and backward directions to simultaneously capture dependencies from past and future contexts and dependencies in the long-term context.

In the CNN model, we design the first layer, which is a linear layer, and the output features are equal to the input features. This linear layer aims to learn a linear mapping that can transform the input data within the same feature space, and this can change the space model to the specific task for BERT embedding. The next build is the standard CNN model. Data flow after passing through the linear layer begins in a convolution layer. It handles the ReLU layer and then applies the max pooling layer to decrease the shape of the data and catch the vital variable in each row. Furthermore, we build four of the same architectures above that propose each can handle and predict a distinct label specified in the dataset as four, and with the three labels can operate by one in four architectures is not used for indicating the output. Concatenate four outputs to a new scalar and use dropout to limit the underfitting and linear to the actual result (the label of the relationship of sentences).

In the BiLSTM model, we take the idea that, like the CNN model above, we build four architectures to present independence for each label. Moreover, before it passes through the layer BiLSTM, we have to add the linear layer because the output of the BERT model has a distinct length, so with this layer, we convert it to the exact shape express generalization for the comparison of each joint model. Also, we have tested it with the BiLSTM and LSTM, but the preeminence of the BiLSTM result, so we eliminated the LSTM and instead used the BiLSTM. Finally, concatenate four outputs of four architecture BiLSTM, apply dropout, and linear to the actual result.

\section{Experiments and Results}
\label{experiment}

\subsection{Datasets Used}
All experiments were performed on the ViNLI dataset. As mentioned above, this experiment checks the quality of this new data set. It needs additional validations to be a typical dataset for Vietnamese NLI. This data set is built and labeled by Vietnamese who have an expert understanding of Vietnamese language and linguistics. The number corpus in the training dataset is 24,376, the dev dataset is 3,009, and the test dataset is 2,991. The dataset is raw from the newspaper so that it can decrease the effect of regional factors and decrease mistakes. This dataset is quite large for Vietnamese now, with 30,376 pairs of premise and hypothesis sentences manually annotated by humans. The statistics on the dataset are shown in Table \ref{table:data}. The difference between this dataset and the standard datasets is that this dataset has an additional label, Other, to distinguish it from the neural label and increase the discriminant of labels.

% Please add the following required packages to your document preamble:
% \usepackage{multirow}
\begin{table}[!ht]
\caption{Quantity of pair in each label in the ViNLI dataset \cite{ViNLI}.}
% \begin{tabular}{|l|llll|}
\centering
\begin{tabular}{ccccc}
\hline
\multirow{2}{*}{\textbf{\textit{Label}}} & \multicolumn{4}{c}{\textbf{\textit{Quantity}}}    \\ \cline{2-5} 
                       & \textbf{\textit{Train}}  & \textbf{\textit{Dev}}   & \textbf{\textit{Test}}  & \textbf{\textit{Total}}  \\ \hline
Entailment             & 6,094  & 739   & 750   & 7,583  \\ \hline
Contradiction          & 6,094  & 764   & 737   & 7,595  \\ \hline
Neutral                & 6,094  & 752   & 777   & 7,623  \\ \hline
Other                  & 6,094  & 754   & 727   & 7,575  \\ \hline
Total                  & 24,376 & 3,009 & 2,991 & 30,376 \\ \hline
\end{tabular}
\label{table:data}
\end{table}

\subsection{Experiment Settings}
We train by NVIDIA A100 GPU through Google Colab and can apply the batch size to 64, except the model XLM-R joint with BiLSTM is large, so the maximum batch size can be set to 32. The BERT model's length parameter equals the max output length after using its tokenizer for all training datasets. Furthermore, we have swapped the order concatenate from premise sequence to hypothesis because the input length in dev or test datasets is longer than the max length above to ensure no missing data. The learning rate used for all experiments is 1e-5, we have tuned this hyperparameter, and 1e-5 is the best situation for all combined models, including BERT or vector space. Besides, we have to check with or without dropout, and the results that should have dropout that can increase 0.4 to 0.5 in accuracy and the same in F1-score

In the neural network, the maximum output size of BERT is 1,024, so in the first linear layer, we change the distinct input size to 1,024. BiLSTM model, we use the layer BiLSTM defined hidden dim to 1,024 and use two layers dimension. In the last linear layer, we drop out 0.1 to prevent overfitting. CNN model, we use different windows for the same purpose.

To implement vector space joint with the neural network, we use three models: fasttext \cite{fasttext}, w2v\_cc\_300d from John Snow LABS\footnote{https://nlp.johnsnowlabs.com/2022/06/23/ner\_living\_species\_fr\_3\_0.html}, and word2-vecVN \cite{word2vecVN}. There are modern models and training in the extensive data. The word2vecVN has various models with different dimensions, but we aim to compare so that we use version 300 dimension. Moreover, we present the output vector model by calculating all vector words' mean. Also, we apply a linear layer that converts the scalar size to 1,024, ensuring that the Neural Network of all experiment have the same architecture.

\subsection{Experiment Results}

Because the problem is that the classification and the output do not have many labels and the test set is enough balance, we use the accuracy and the macro F1-score for benchmark which is simple and convenient, and benchmark in three labels and four labels. However, as we can see in Table \ref{table:data}, the data is almost balanced, and to increase objectivity and credibility, we use accuracy as an additional evaluation metric. Based on the research \cite{Phobert_CNN}, we use Equation \eqref{eq:accuracy} to Equation \eqref{eq:F1} presents the metric measure for this study. Where $i \in \{1, 2, 3\}$ (denoted by entailment, contradiction, neutral) and $\text{tp}_{i}$, $\text{fp}_{i}$, $\text{tn}_{i}$, $\text{fn}_{i}$ sequentially true positive, false positive, true negative, and false negative. Moreover, $\text{M} $ represents macro-averaging value. The results are shown in Table \ref{table:result} with the experimental results \cite{ViNLI}. Moreover, in the table, the result gains better performance in bold.

% Please add the following required packages to your document preamble:
% \usepackage{multirow}
% \usepackage[table,xcdraw]{xcolor}
% If you use beamer only pass "xcolor=table" option, i.e. \documentclass[xcolor=table]{beamer}
\begin{table}[ht]
\caption{Experimental result of joint models for NLI.}
\centering
\resizebox{\columnwidth}{!}{
\begin{tabular}{ccllllllll}
\hline
\multicolumn{1}{c}{}                         &                                  & \multicolumn{4}{c}{Three label}                                                                                                                                                                                                     & \multicolumn{4}{c}{Four label}                                                                                                                                                                                                     \\ \cline{3-10} 
\multicolumn{1}{c}{}                         &                                  & \multicolumn{2}{c}{Dev}                                                                                                  & \multicolumn{2}{c}{Test}                                                                            & \multicolumn{2}{c}{Dev}                                                                                                  & \multicolumn{2}{c}{Test}                                                                            \\ \cline{3-10} 
\multicolumn{1}{c}{\multirow{-3}{*}{Model}}  & \multirow{-3}{*}{Word embedding} & \multicolumn{1}{c}{Acc}                                    & \multicolumn{1}{c}{F1}                                     & \multicolumn{1}{c}{Acc}                                    & \multicolumn{1}{c}{F1}                & \multicolumn{1}{c}{Acc}                                    & \multicolumn{1}{c}{F1}                                     & \multicolumn{1}{c}{Acc}                                    & \multicolumn{1}{c}{F1}                \\ \hline
\multicolumn{2}{c}{PhoBERT (Huynh et al. (2022) \cite{ViNLI}}                                 & \multicolumn{1}{l}{{\color[HTML]{000000} 0.7733}}          & \multicolumn{1}{l}{{\color[HTML]{000000} 0.7734}}          & \multicolumn{1}{l}{{\color[HTML]{000000} 0.7593}}          & {\color[HTML]{000000} 0.7587}          & \multicolumn{1}{l}{{\color[HTML]{000000} 0.8072}}          & \multicolumn{1}{l}{{\color[HTML]{000000} 0.8072}}          & \multicolumn{1}{l}{{\color[HTML]{000000} 0.8067}}          & {\color[HTML]{000000} 0.8069}          \\ \hline
\multicolumn{2}{c}{XLM-R (Huynh et al. (2022) \cite{ViNLI}}                                   & \multicolumn{1}{l}{{\color[HTML]{000000} 0.8302}}          & \multicolumn{1}{l}{{\color[HTML]{000000} 0.8298}}          & \multicolumn{1}{l}{{\color[HTML]{000000} 0.8136}}          & {\color[HTML]{000000} 0.8131}          & \multicolumn{1}{l}{{\color[HTML]{000000} 0.8677}}          & \multicolumn{1}{l}{{\color[HTML]{000000} 0.8676}}          & \multicolumn{1}{l}{{\color[HTML]{000000} 0.8599}}          & {\color[HTML]{000000} 0.8610}           \\ \hline
\multicolumn{2}{c}{mBERT (Huynh et al. (2022) \cite{ViNLI}}                                   & \multicolumn{1}{l}{{\color[HTML]{000000} 0.6741}}          & \multicolumn{1}{l}{{\color[HTML]{000000} 0.6746}}          & \multicolumn{1}{l}{{\color[HTML]{000000} 0.6484}}          & {\color[HTML]{000000} 0.6483}          & \multicolumn{1}{l}{{\color[HTML]{000000} 0.7391}}          & \multicolumn{1}{l}{{\color[HTML]{000000} 0.7383}}          & \multicolumn{1}{l}{{\color[HTML]{000000} 0.7345}}          & {\color[HTML]{000000} 0.7362}          \\ \hline
\multicolumn{1}{c}{}                         & PhoBERT                          & \multicolumn{1}{l}{{\color[HTML]{000000} 0.7694}}          & \multicolumn{1}{l}{{\color[HTML]{000000} 0.7694}}          & \multicolumn{1}{l}{{\color[HTML]{000000} \textbf{0.7597}}} & {\color[HTML]{000000} \textbf{0.7595}} & \multicolumn{1}{l}{{\color[HTML]{000000} \textbf{0.8169}}} & \multicolumn{1}{l}{{\color[HTML]{000000} \textbf{0.8168}}} & \multicolumn{1}{l}{{\color[HTML]{000000} \textbf{0.8091}}} & {\color[HTML]{000000} \textbf{0.8099}} \\ \cline{2-10} 
\multicolumn{1}{c}{}                         & XLM-R                            & \multicolumn{1}{l}{{\color[HTML]{000000} \textbf{0.8395}}} & \multicolumn{1}{l}{{\color[HTML]{000000} \textbf{0.8392}}} & \multicolumn{1}{l}{{\color[HTML]{000000} \textbf{0.8277}}} & {\color[HTML]{000000} \textbf{0.8278}} & \multicolumn{1}{l}{{\color[HTML]{000000} 0.8637}}          & \multicolumn{1}{l}{{\color[HTML]{000000} 0.8634}}          & \multicolumn{1}{l}{{\color[HTML]{000000} 0.8579}}          & {\color[HTML]{000000} 0.8587}          \\ \cline{2-10} 
\multicolumn{1}{c}{}                         & mBERT                            & \multicolumn{1}{l}{{\color[HTML]{000000} 0.6483}}          & \multicolumn{1}{l}{{\color[HTML]{000000} 0.6481}}          & \multicolumn{1}{l}{{\color[HTML]{000000} 0.6400}}            & {\color[HTML]{000000} 0.6392}          & \multicolumn{1}{l}{{\color[HTML]{000000} 0.7215}}          & \multicolumn{1}{l}{{\color[HTML]{000000} 0.7210}}           & \multicolumn{1}{l}{{\color[HTML]{000000} 0.7212}}          & {\color[HTML]{000000} 0.7225}          \\ \cline{2-10} 
\multicolumn{1}{c}{\multirow{-4}{*}{CNN}}    & InfoXLM                          & \multicolumn{1}{l}{{\color[HTML]{000000} 0.8293}}          & \multicolumn{1}{l}{{\color[HTML]{000000} 0.8293}}          & \multicolumn{1}{l}{{\color[HTML]{000000} 0.8154}}          & {\color[HTML]{000000} 0.8157}          & \multicolumn{1}{l}{{\color[HTML]{000000} 0.8671}}          & \multicolumn{1}{l}{{\color[HTML]{000000} 0.8674}}          & \multicolumn{1}{l}{{\color[HTML]{000000} 0.8589}}          & {\color[HTML]{000000} 0.8606}          \\ \hline
\multicolumn{1}{c}{}                         & PhoBERT                          & \multicolumn{1}{l}{{\color[HTML]{000000} 0.7503}}          & \multicolumn{1}{l}{{\color[HTML]{000000} 0.7502}}          & \multicolumn{1}{l}{{\color[HTML]{000000} 0.7398}}          & {\color[HTML]{000000} 0.7384}          & \multicolumn{1}{l}{{\color[HTML]{000000} 0.7940}}           & \multicolumn{1}{l}{{\color[HTML]{000000} 0.7936}}          & \multicolumn{1}{l}{{\color[HTML]{000000} 0.7837}}          & {\color[HTML]{000000} 0.7840}           \\ \cline{2-10} 
\multicolumn{1}{c}{}                         & XLM-R                            & \multicolumn{1}{l}{{\color[HTML]{000000} 0.8262}}          & \multicolumn{1}{l}{{\color[HTML]{000000} 0.8262}}          & \multicolumn{1}{l}{{\color[HTML]{000000} \textbf{0.8255}}} & {\color[HTML]{000000} \textbf{0.8255}} & \multicolumn{1}{l}{{\color[HTML]{000000} 0.8598}}          & \multicolumn{1}{l}{{\color[HTML]{000000} 0.8594}}          & \multicolumn{1}{l}{{\color[HTML]{000000} 0.8455}}          & {\color[HTML]{000000} 0.8466}          \\ \cline{2-10} 
\multicolumn{1}{c}{}                         & mBERT                            & \multicolumn{1}{l}{{\color[HTML]{000000} 0.6608}}          & \multicolumn{1}{l}{{\color[HTML]{000000} 0.6599}}          & \multicolumn{1}{l}{{\color[HTML]{000000} \textbf{0.6612}}} & {\color[HTML]{000000} \textbf{0.6578}} & \multicolumn{1}{l}{{\color[HTML]{000000} 0.7305}}          & \multicolumn{1}{l}{{\color[HTML]{000000} 0.7306}}          & \multicolumn{1}{l}{{\color[HTML]{000000} 0.7155}}          & {\color[HTML]{000000} 0.7179}          \\ \cline{2-10} 
\multicolumn{1}{c}{\multirow{-4}{*}{BiLSTM}} & InfoXLM                          & \multicolumn{1}{l}{{\color[HTML]{000000} 0.8333}}          & \multicolumn{1}{l}{{\color[HTML]{000000} 0.8332}}          & \multicolumn{1}{l}{{\color[HTML]{000000} 0.8228}}          & {\color[HTML]{000000} 0.8233}          & \multicolumn{1}{l}{{\color[HTML]{000000} 0.8647}}          & \multicolumn{1}{l}{{\color[HTML]{000000} 0.8649}}          & \multicolumn{1}{l}{{\color[HTML]{000000} 0.8592}}          & {\color[HTML]{000000} 0.8607}          \\ \hline
\end{tabular}}
\label{table:result}
\end{table}

\begin{equation}\label{eq:accuracy}
\text{Accuracy} = \frac{\sum_{i=1}^{3}\frac{tp_{i}+tn_{i}}{tp_{i}+fp_{i}+tn_{i}+fn_{i}}}{3}
\end{equation}

\begin{equation}\label{eq:precision}
\text{Precision}_{M} = \frac{\sum_{i=1}^{3}\frac{tp_{i}}{tp_{i}+fp_{i}}}{3}
\end{equation}

\begin{equation}\label{eq:recall}
\text{Recall}_{M} = \frac{\sum_{i=1}^{3}\frac{tp_{i}}{tp_{i}+fn_{i}}}{3}
\end{equation}

\begin{equation}\label{eq:F1}
\text{F1}_{M} = 2 * \frac{\text{Precision}_{M} * \text{Recall}_{M}}{\text{Precision}_{M} + \text{Recall}_{M}}
\end{equation}

\section{Result Analysis}
\label{result}

%\subsection{F1 score}

As the results table shows, the results differ in the range of values $\pm2.3\%$ between this approach and Huynh et al.'s approach, including the development and test sets. Furthermore, this approach does not outperform the previous experiment, but it has several better points that we record.

In the performance comparison, the model that has XLM-R in part consistently outperformed PhoBERT and mBERT across all label settings, and this seems obvious because of its size and complex architecture. Furthermore, CNN models generally performed better than BiLSTM models across different word embeddings, and this suggests that CNN architectures are more effective at extracting meaningful features for the task of word embedding classification. Moreover, this is also easy to see because the CNN architecture aims to extract features from data present in the matrix form. On the other hand, BiLSTM is an architecture for a sequence-to-sequence data. In a study comparing non-contextualized and contextualized vectors in Table \ref{table:compare}, we record the result of contextualized outperformance of the non-contextualized vector. In several cases, the results are more than twice as good. 

% Please add the following required packages to your document preamble:
% \usepackage{multirow}
% \usepackage[table,xcdraw]{xcolor}
% If you use beamer only pass "xcolor=table" option, i.e. \documentclass[xcolor=table]{beamer}
\begin{table}[ht]
\caption{Comparing the vector space with BERT about generating the embedding in accuracy (\%).}
\centering
% \resizebox{\columnwidth}{!}{
% \begin{tabular}{cccrrrr}
\begin{tabular}{cccccc}
% \hline
\cline{1-6}
\multirow{2}{*}{\textbf{Model}}  & \multirow{2}{*}{\textbf{Word embedding}} & \multicolumn{2}{c}{\textbf{Three labels}} & \multicolumn{2}{c}{\textbf{Four labels}} \\ \cline{3-6} 
                        &                                 & \textbf{Dev}          & \textbf{Test}         & \textbf{Dev}          & \textbf{Test}         \\ \cline{1-6} %\hline
\multirow{7}{*}{CNN}    & word2vecVN                      & 41.82        & 42.05        & 35.89        & 35.54        \\  
                        & fasttext                        & 39.11        & 38.16        & 34.80        & 31.83        \\  
                        & w2v\_cc\_300d                   & 36.85        & 37.99        & 33.80        & 31.29        \\  
                        & PhoBERT                         & 76.94        & 75.97        & 81.69        & 80.91        \\  
                        & XLM-R                           & 83.95        & 82.77        & 86.37        & 85.79        \\  
                        & mBERT                           & 64.83        & 64.40        & 72.15        & 72.12        \\  
                        & InfoXLM                         & 82.93        & 81.54        & 86.71        & 85.89        \\ \cline{1-6} %\hline
\multirow{7}{*}{BiLSTM} & word2vecVN                      & 42.26        & 42.58        & 39.75        & 39.28        \\  
                        & fasttext                        & 39.02        & 37.23        & 34.06        & 32.53        \\  
                        & w2v\_cc\_300d                   & 37.92        & 37.59        & 34.06        & 32.53        \\  
                        & PhoBERT                         & 75.03        & 73.98        & 79.40        & 78.37        \\  
                        & XLM-R                           & 82.62        & 82.55        & 85.98        & 84.55        \\  
                        & mBERT                           & 66.08        & 66.12        & 73.05        & 71.55        \\  
                        & InfoXLM                         & 83.33        & 82.28        & 86.47        & 85.92        \\ \hline
\end{tabular}
\label{table:compare}
\end{table}

Compared with the previous experiment, in the dev dataset, most studies have results approximately to previous studies, but the results still need to be improved. However, the PhoBERT - CNN model and XLM-R - BiLSTM results improve by nearly 1\%.
In the test dataset, the PhoBERT joined with CNN gave little better results than PhoBERT in three and four labels. Moreover, in the XLM-R model, the combination with CNN or BiLSTM, whose results are remarkable in the standard three-label dataset when it increases (+1\%), the same goes for mBERT on the same dataset. However, its results do not improve in the four labels dataset, both XLM-R and mBERT combined with CNN or BiLSTM.
Besides, almost the combination of CLM with CNN gives better results than with BiLSTM, and this is easy to understand because the context output of CLM in this experiment is a matrix. Nevertheless, we also claim that if we change the input of the Neural Network to a single context CLM, combining it with BiLSTM is better and the opposite of mBERT.

%\tab\textbf{In Accuracy}
The XLM-R model, particularly as reported by Huynh et al. (2022) \cite{ViNLI}, served as a strong benchmark with impressive accuracies. However, our experiments with standalone PhoBERT and XLM-R models yielded consistently robust performances, with XLM-R surpassing the benchmarks on the three-label task, highlighting the model's ability to generalize effectively.

The mBERT model, under Tin et al.'s configuration \cite{ViNLI}, showed lower accuracy, which was also reflected in our standalone trials. This consistency across different studies underscores the model's limitations for the given tasks. On the other hand, the use of InfoXLM as an embedding in our CNN and BiLSTM models significantly improved their performance, suggesting that the right combination of model architecture and word embeddings is pivotal for optimizing accuracy.

For the four-label classification, our findings were congruent with Tin et al.'s work \cite{ViNLI} in that the complexity of the task slightly reduced model accuracies. Nevertheless, the BiLSTM with InfoXLM embedding from our experiments demonstrated a noteworthy resilience to this increased complexity, maintaining high accuracy and underscoring the robustness of this joint model.

In essence, this study not only validates the findings of Tin et al. (2022) \cite{ViNLI} but also expands the landscape of possibilities for future explorations in natural language processing. Our results advocate for a nuanced approach to model selection, where the specifics of the dataset and task at hand are matched with an appropriate word embedding to yield optimal performance.

We have analyzed the linguist aspect that affects several models for a more general view. The first is XLM-R combined with CNN, which has the highest score. Furthermore, we analyze the mBERT model, which has the worst result, and PhoBERT, the model is trained only for the Vietnamese language, and both combine with BiLSTM to be able to analyze the impact of BiLSTM. We have studied the questions raised for the linguist aspect of models following Huynh et al. (2022) \cite{ViNLI} in the three-label dataset.

\textbf{Comparing with ChatGPT} \\
We have random sampling in the test set and take one hundred pairs to compare between ChatGPT\footnote{https://chat.openai.com/ and evaluate model GPT-3.5 on 9/24/2023} and the joint model. We utilize a one-shot learning technique, where the input is prompted with the hypothesis and premise pair. Further details regarding the input prompting can be found in the Appendix section. We archive twice as a result of almost joint models as ChatGPT, detailed in Table \ref{table:chat-gpt}.

\begin{table}[ht]
\caption{Comparing performance generated by ChatGPT with performance BERT about Vietnamese natural language inference.}
\centering
\begin{tabular}{cccc}
\hline
\textbf{Model}           & \textbf{Word embeding} & \textbf{Acc}                                        & \textbf{F1} \\ 
% \hline 
\cline{1-4}
\multicolumn{2}{c}{ChatGPT}                       & 0.42                                                & 0.38      \\ 
% \hline
\cline{1-4}
                         & PhoBERT                & 0.80                                                 & 0.80      \\  
                         & XLM-R                  & 0.83                                                & 0.84      \\  
                         & mBERT                  & 0.69                                                & 0.71      \\  
\multirow{-4}{*}{CNN}    & InfoXLM                & 0.89                                                & 0.89      \\ 
% \hline
\cline{1-4}
                         & PhoBERT                & 0.65                                                & 0.66      \\  
                         & XLM-R                  & 0.84                                                & 0.85      \\ 
                         & mBERT                  & 0.69                                                & 0.70      \\  
\multirow{-4}{*}{BiLSTM} & InfoXLM                & 0.87 & 0.87      \\ \hline
\end{tabular}
\label{table:chat-gpt}
\end{table}
We conducted inference using the ChatGPT through API calls. OpenAI has not publicly disclosed the specific amount of VRAM required for the inference process, given the hardware limitations associated with ChatGPT's extensive 175 billion parameters \cite{chatgpt} In our experimental evaluations, we observed distinct VRAM needs for loading different language models. Specifically, loading of XLM-R large required 2GB of VRAM, mBERT required 837MB, and PhoBERT demanded VRAM capacities of 1529MB with model sizes of \textasciitilde560M, \textasciitilde178M, and \textasciitilde369M parameters, respectively. This diversity in VRAM requirements highlights the efficacy and superiority of our proposed method.

\textbf{Predicted Performance for Each Inference Label}.\\
Table \ref{table:accuracy} shows the performance of the model on each label in the validation set on three labels of models that we have noted above. We calculate the accuracy of each label to exploit the disadvantages and advantages of specific problems. PhoBERT and XLM-R predict well in the Entailment label, and mBERT indicates well in the Neutral label. The difference in accuracy between the highest and lowest labels is most significant in mBERT compared to PhoBERT and least in XLM-R, which also holds for the model's performance.

% Please add the following required packages to your document preamble:
% \usepackage{multirow}
\begin{table}[htbp]
\caption{Model performance per label.}
\centering
\begin{tabular}{lccc}
\hline
\multirow{2}{*}{\textbf{Labels}} & \multicolumn{2}{c}{\textbf{BiLSTM}} & \textbf{CNN}   \\ \cline{2-4} 
                        & PhoBERT       & mBERT      & XLM-R \\ \hline
Entailment              & 80.51         & 65.49      & 87.01 \\
Contradiction           & 70.29         & 59.69      & 85.47 \\
Neutral                 & 74.47         & 73.13      & 79.39 \\ \hline
\end{tabular}
\label{table:accuracy}
\end{table}

\textbf{Predictive performance for each topic}.\\
Table \ref{table:topic} shows each topic's model performance to observe the open-domain's impact and compare it with the previous study. The accuracy between the topics for each model does not significantly differ. It is easy to recognize that the model with better performance excels in accuracy for each topic.

% Please add the following required packages to your document preamble:
% \usepackage{multirow}
\begin{table}[ht]
\caption{Model performance per topic.}
\centering
\begin{tabular}{lccc}
% \hline
\cline{1-4}
\multirow{2}{*}{\textbf{Labels}} & \multicolumn{2}{c}{\textbf{BiLSTM}} & \textbf{CNN}   \\ \cline{2-4} 
                        & \textbf{PhoBERT}       & \textbf{mBERT}      & \textbf{XLM-R} \\ \hline
Technology              & 71.10          & 63.01      & 84.97 \\
Tourism                 & 77.91         & 64.53      & 87.21 \\
Education               & 76.57         & 69.14      & 84.00  \\
Entertainment           & 80.57         & 68.00       & 86.86 \\
Science                 & 74.27         & 67.25      & 85.38 \\
Business                & 73.26         & 65.12      & 80.23 \\
Law                     & 76.57         & 65.14      & 80.57 \\
Health                  & 84.39         & 73.41      & 85.55 \\
World                   & 70.69         & 61.49      & 81.61 \\
Sports                  & 71.26         & 62.07      & 81.61 \\
News                    & 73.41         & 75.72      & 85.55 \\
Vechicles               & 72.41         & 62.64      & 86.21 \\
Life                    & 72.99         & 61.49      & 81.61 \\ \hline
\end{tabular}
\label{table:topic}
\end{table}

\begin{figure}[htbp]
    \centering
    \begin{subfigure}{\textwidth} % Adjust the width as needed
        \centering
        \includegraphics[width=7cm]
{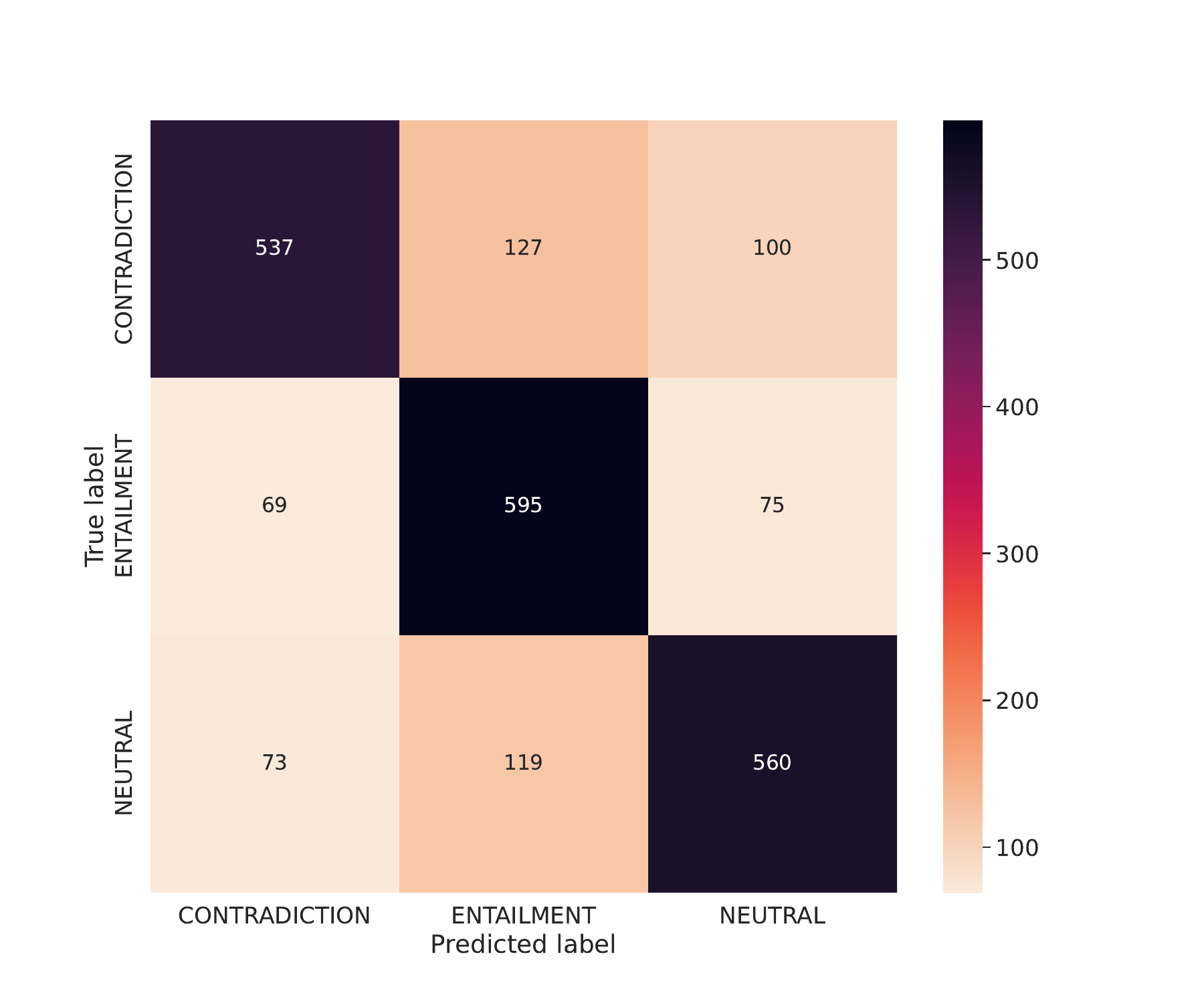}
        \caption{PhoBERT - BiLSTM confusion matrix.}
        \label{pic:PhoBERT - BiLSTM}
    \end{subfigure}%
    \hfill % Add horizontal space between subfigures
    \begin{subfigure}{\textwidth}
        \centering
        \includegraphics[width=7cm]{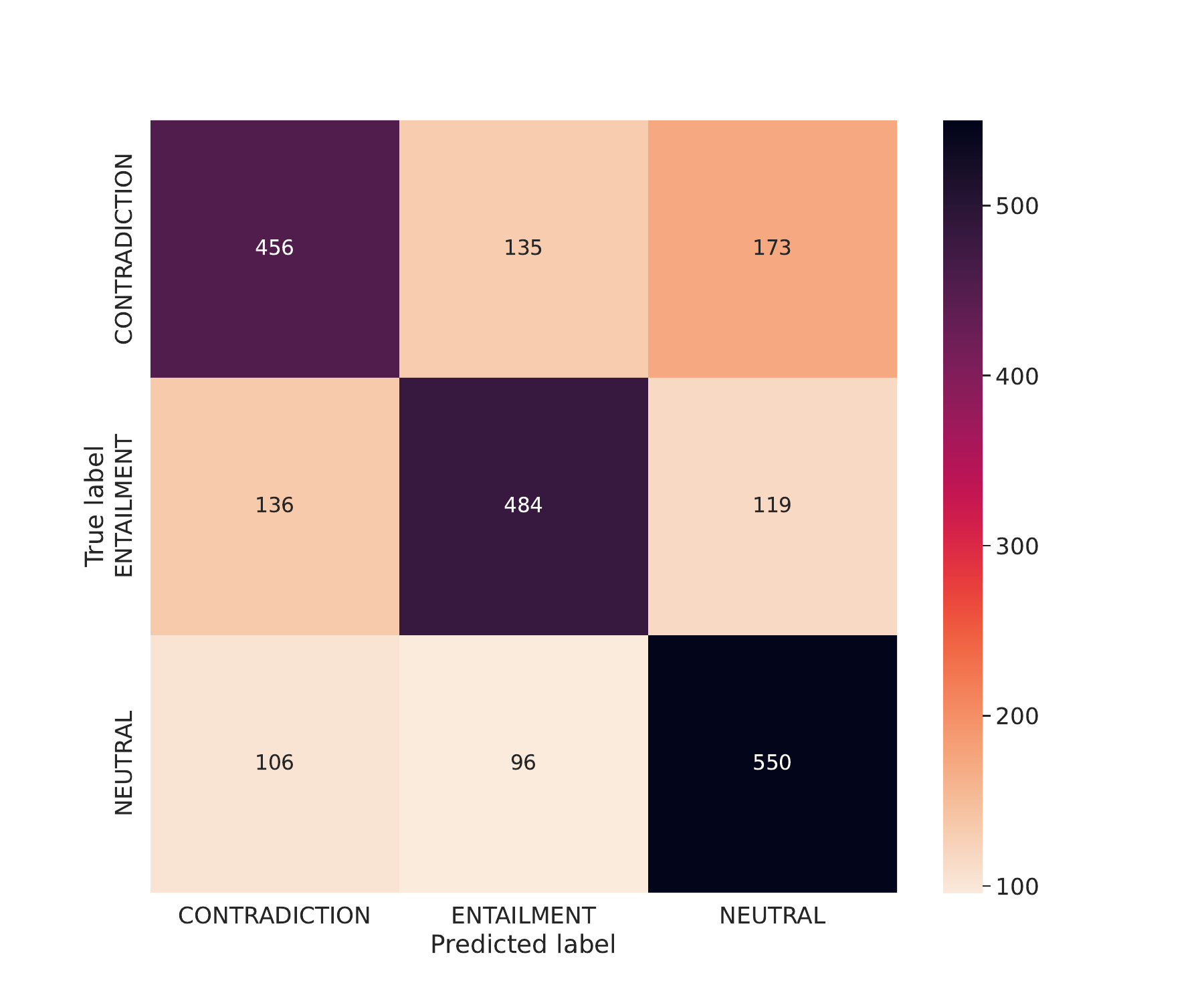}
        \caption{mBERT - BiLSTM confusion matrix.}
        \label{pic:mBERT - BiLSTM}
    \end{subfigure}%
    \hfill % Add horizontal space between subfigures
    % \vspace{50pt}
    \begin{subfigure}{\textwidth}
        \centering
        \includegraphics[width=7cm]{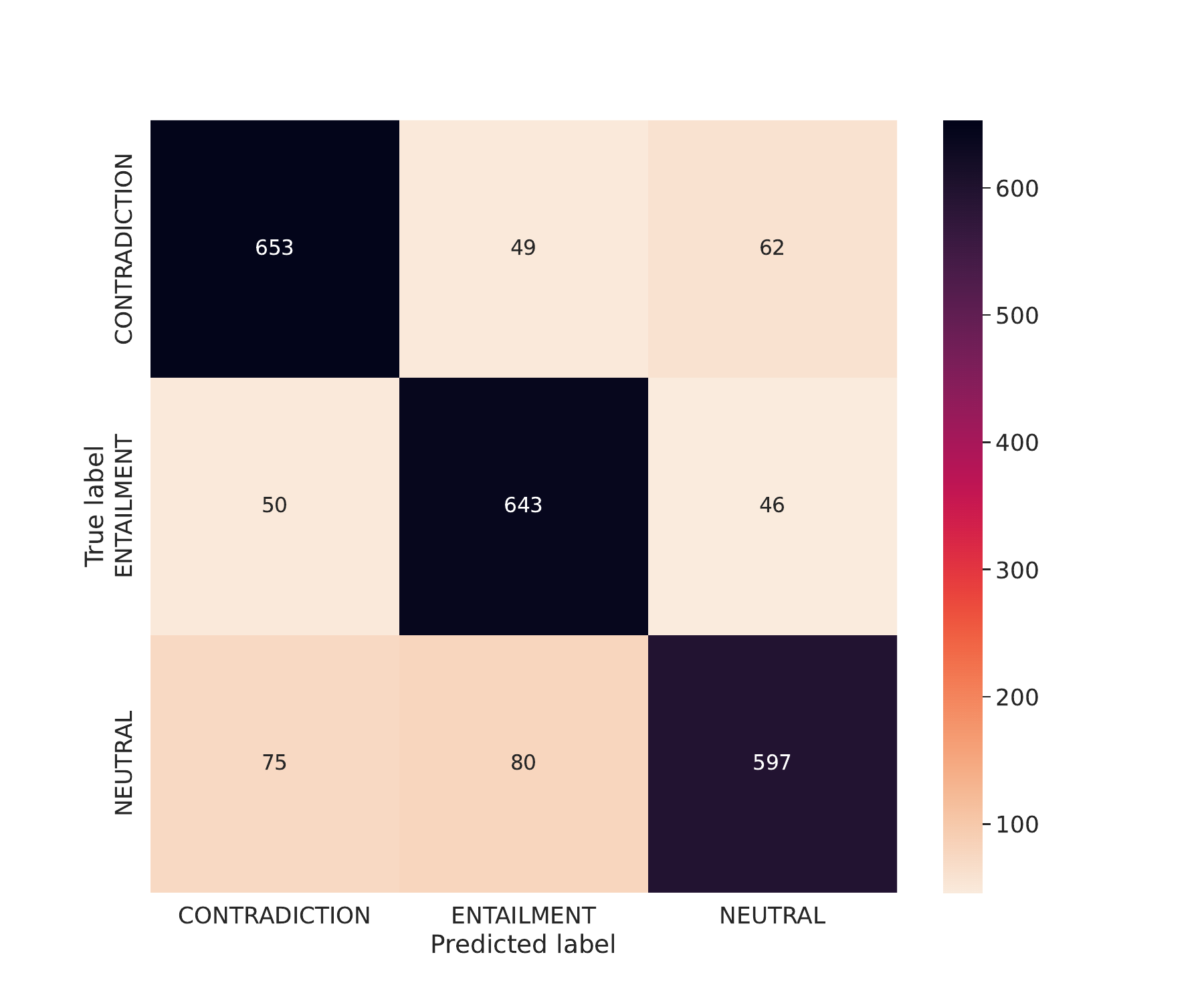}
        \caption{XLM-R - CNN confusion matrix.}
        \label{pic:XLM-R - CNN}
    \end{subfigure}
    \caption{Three Image Sequences}
    \label{pic:confussion}
    \label{fig:Confusion matrix of differ joint model}
\end{figure}

\textbf{Confusion Matrix}\\
We investigate the confusion matrix and present several fascinating results in Fig. \ref{pic:confussion}. The Contradiction actual label, predicted worst in the PhoBERT-BiLSTM model, is similarly accurate in the mBERT-BiLSTM. Additionally, the mBERT model makes a considerably erroneous prediction from the Neutral label to the CONTRADICTION label, much like in the prior experiment cited in ViNLI. The XLM-R-CNN model also provides the best forecast across all labels, and the label deviation is also reasonable.

\textbf{Data Sampling}
We perform data sampling and test how each of the models predicts labels. We obtained many interesting samples, Table \ref{summary}. Huynh et al.'s model fail at the same time in all case when the gold label has a 'contradiction' label, and it always predicts the 'neutral' label. Our approach can predict that case in several joint model. In the final case, virtually almost all joint models (7/8 cases) can correctly predict that the correct label is "contradiction," and all language models wrongly predicted to be labeled "neutral." On top of that, it also shows that the dataset still has a lot of challenges to solve.

% Please add the following required packages to your document preamble:
% \usepackage{multirow}
% \usepackage{graphicx}
\begin{table}[h]
\caption{Sample classifications for ViNLI test set.}
\resizebox{\textwidth}{!}{%
\begin{tabular}{llccc}
\hline
\multicolumn{1}{c}{\textbf{Hypothesis}}                                                                                                                                                                                  & \multicolumn{1}{c}{\textbf{Premise}}                                                                                                         & \textbf{Gold label}             & \textbf{Model}   & \textbf{Predict class} \\ \hline
\multirow{10}{4cm}{Nạn nhân không bị nguy hiểm đến tính mạng nhưng chưa thể làm việc với cơ quan điều tra. \textbf{\textit{(The victim is not in immediate danger of death but has not been able to work with the authorities.)}}}                                                                                                                & \multirow{10}{4cm}{Cơ quan điều tra không thể làm việc với nạn nhân vì sau vụ án nạn nhân bị tử vong. \textbf{\textit{(The coroner couldn't work with the victim because after the case the victim died.)}}}                                         & \multirow{10}{1.5cm}{contradiction} & PhoBERT - CNN    & neutral                \\
                                                                                                                                                                                                                         &                                                                                                                                              &                                 & PhoBERT - BiLSTM & neutral                \\
                                                                                                                                                                                                                         &                                                                                                                                              &                                 & XLM-R - CNN      & contradiction          \\
                                                                                                                                                                                                                         &                                                                                                                                              &                                 & XLM-R - BiLSTM   & neutral                \\
                                                                                                                                                                                                                         &                                                                                                                                              &                                 & mBERT - CNN      & neutral                \\
                                                                                                                                                                                                                         &                                                                                                                                              &                                 & mBERT - BiLSTM   & neutral                \\
                                                                                                                                                                                                                         &                                                                                                                                              &                                 & InfoXLM - CNN    & neutral                \\
                                                                                                                                                                                                                         &                                                                                                                                              &                                 & InfoXLM - BiLSTM & neutral                \\
                                                                                                                                                                                                                         &                                                                                                                                              &                                 & PhoBERT          & neutral                \\
                                                                                                                                                                                                                         &                                                                                                                                              &                                 & XLM-R            & neutral                \\
                                                                                                                                                                                                                         &                                                                                                                                              &                                 & mBERT            & neutral                \\ \hline
\multirow{10}{4cm}{Giá các mặt hàng dầu đều tăng. \textbf{\textit{(Prices of oil commodities are rising.)}}}                                                                                                                                                                         & \multirow{10}{4cm}{Bên cạnh một số mặt hàng dầu trong nước đang tăng giá, thì một số mặt hàng dầu được nhập khẩu từ một số nước thì lại giảm. \textbf{\textit{(Besides several domestic oil commodities that are rising in price, several oil commodities imported from several countries are falling.)}}} & \multirow{10}{1.5cm}{contradiction} & PhoBERT - CNN    & neutral                \\
                                                                                                                                                                                                                         &                                                                                                                                              &                                 & PhoBERT - BiLSTM & neutral                \\
                                                                                                                                                                                                                         &                                                                                                                                              &                                 & XLM-R - CNN      & neutral          \\
                                                                                                                                                                                                                         &                                                                                                                                              &                                 & XLM-R - BiLSTM   & neutral                \\
                                                                                                                                                                                                                         &                                                                                                                                              &                                 & mBERT - CNN      & contradiction                \\
                                                                                                                                                                                                                         &                                                                                                                                              &                                 & mBERT - BiLSTM   & neutral                \\
                                                                                                                                                                                                                         &                                                                                                                                              &                                 & InfoXLM - CNN    & neutral                \\
                                                                                                                                                                                                                         &                                                                                                                                              &                                 & InfoXLM - BiLSTM & neutral                \\
                                                                                                                                                                                                                         &                                                                                                                                              &                                 & PhoBERT          & neutral                \\
                                                                                                                                                                                                                         &                                                                                                                                              &                                 & XLM-R            & neutral                \\
                                                                                                                                                                                                                         &                                                                                                                                              &                                 & mBERT            & neutral                \\ \hline
\multirow{10}{4cm}{Với cách biệt hai điểm, Barca sẽ soán ngôi đầu nếu hạ Atletico. \textbf{\textit{(With a two - point gap, Barca would take the lead if they defeated Atletico.)}}}                                                                                                                                        & \multirow{10}{4cm}{Chiến thắng dường như đã được dự đoán thuộc về Atletico. \textbf{\textit{(The victory seemed to have been predicted to belong to Atletico.)}}}                                                                   & \multirow{10}{1.5cm}{contradiction} & PhoBERT - CNN    & contradiction                \\
                                                                                                                                                                                                                         &                                                                                                                                              &                                 & PhoBERT - BiLSTM & neutral                \\
                                                                                                                                                                                                                         &                                                                                                                                              &                                 & XLM-R - CNN      & neutral          \\
                                                                                                                                                                                                                         &                                                                                                                                              &                                 & XLM-R - BiLSTM   & neutral                \\
                                                                                                                                                                                                                         &                                                                                                                                              &                                 & mBERT - CNN      & contradiction                \\
                                                                                                                                                                                                                         &                                                                                                                                              &                                 & mBERT - BiLSTM   & contradiction                \\
                                                                                                                                                                                                                         &                                                                                                                                              &                                 & InfoXLM - CNN    & neutral                \\
                                                                                                                                                                                                                         &                                                                                                                                              &                                 & InfoXLM - BiLSTM & contradiction                \\
                                                                                                                                                                                                                         &                                                                                                                                              &                                 & PhoBERT          & neutral                \\
                                                                                                                                                                                                                         &                                                                                                                                              &                                 & XLM-R            & neutral                \\
                                                                                                                                                                                                                         &                                                                                                                                              &                                 & mBERT            & entailment                \\ \hline
\multirow{10}{4cm}{Trong lần gặp lại này, Zverev vượt trội đối thủ ở giao bóng. \textbf{\textit{(In this rematch, Zverev outperformed his opponent in the serve.)}}}                                                                                                                                           & \multirow{10}{4cm}{Đối thủ có kỹ năng giao bóng vượt xa Zverev.\textbf{\textit{(Opponents have better passing skills than Zverev.)}}}                                                                               & \multirow{10}{1.5cm}{contradiction} & PhoBERT - CNN    & entailment                \\
                                                                                                                                                                                                                         &                                                                                                                                              &                                 & PhoBERT - BiLSTM & entailment                \\
                                                                                                                                                                                                                         &                                                                                                                                              &                                 & XLM-R - CNN      & contradiction          \\
                                                                                                                                                                                                                         &                                                                                                                                              &                                 & XLM-R - BiLSTM   & contradiction                \\
                                                                                                                                                                                                                         &                                                                                                                                              &                                 & mBERT - CNN      & entailment                \\
                                                                                                                                                                                                                         &                                                                                                                                              &                                 & mBERT - BiLSTM   & entailment                \\
                                                                                                                                                                                                                         &                                                                                                                                              &                                 & InfoXLM - CNN    & entailment                \\
                                                                                                                                                                                                                         &                                                                                                                                              &                                 & InfoXLM - BiLSTM & entailment                \\
                                                                                                                                                                                                                         &                                                                                                                                              &                                 & PhoBERT          & entailment                \\
                                                                                                                                                                                                                         &                                                                                                                                              &                                 & XLM-R            & entailment                \\
                                                                                                                                                                                                                         &                                                                                                                                              &                                 & mBERT            & entailment                \\ \hline
\multirow{10}{4cm}{Lý Chấn Cường, 41 tuổi, dùng kéo tấn công tài xế, cướp taxi tại quận 1 rồi chạy hơn 100 km đâm vào nhà dân trước khi bị cảnh sát bắt giữ. \textbf{\textit{(Ly Chan Cuong, 41, assaulted a driver with scissors, robbed a taxi in District 1 and then ran more than 100 kilometers into a resident's house before being arrested by police.)}}}                                                              & \multirow{10}{4cm}{Lý Chân Cường dùng kéo uy hiếp nhưng đã bị người tài xế khống chế khiến hắn bị bắt trong phi vụ cướp bất thành. \textbf{\textit{(Ly Chan Cuong used scissors to intimidate him, but was overpowered by a driver who caught him in a botched robbery attempt.)}}}            & \multirow{10}{1.5cm}{contradiction} & PhoBERT - CNN    & contradiction                \\
                                                                                                                                                                                                                         &                                                                                                                                              &                                 & PhoBERT - BiLSTM & contradiction                \\
                                                                                                                                                                                                                         &                                                                                                                                              &                                 & XLM-R - CNN      & neutral          \\
                                                                                                                                                                                                                         &                                                                                                                                              &                                 & XLM-R - BiLSTM   & contradiction                \\
                                                                                                                                                                                                                         &                                                                                                                                              &                                 & mBERT - CNN      & contradiction                \\
                                                                                                                                                                                                                         &                                                                                                                                              &                                 & mBERT - BiLSTM   & contradiction                \\
                                                                                                                                                                                                                         &                                                                                                                                              &                                 & InfoXLM - CNN    & contradiction                \\
                                                                                                                                                                                                                         &                                                                                                                                              &                                 & InfoXLM - BiLSTM & contradiction                \\
                                                                                                                                                                                                                         &                                                                                                                                              &                                 & PhoBERT          & neutral                \\
                                                                                                                                                                                                                         &                                                                                                                                              &                                 & XLM-R            & neutral                \\
                                                                                                                                                                                                                         &                                                                                                                                              &                                 & mBERT            & neutral                \\ \\ \hline
\end{tabular}%
}
\label{summary}
\end{table}

Overall, the F1-score was consistently high across different models, suggesting that the models can understand the language and are not overfitting to the training dataset, given that the joint BERT embedding and neural network is a good approach. The contextualized vector makes paragraph embedding better than the non-contextualized vector in all cases. It understands the context and makes the connection between the words in the sentence more closely. Furthermore, by experimenting with chatGPT, we archive the Vietnamese NLI task, which is a tricky task; it needs more than the one-shot learning technique. In addition, the data set still has many challenges to evaluate and solve. Moreover, each model has different advantages and disadvantages, so be careful when choosing any model. Therefore, we should try with many entries to obtain the best combination model that fits in an NLP task.

\section{Conclusion and Future Work}
\label{conclusion}

One-shot learning is not good performance in Vietnamese NLI tasks. The contextualized vector outperforms the non-contextualized vector by up to twice the performance. CNN and BiLSTM are ineffective at encoding contextual information in both pre- and post-context in the Vietnamese language. Therefore, the use of a contextual language model helps to improve the contextual awareness of the models, improving the efficiency of CNN and BiLSTM. Therefore, a joint model represents a promising approach that leverages BERT's comprehension for contextual sentences and employs artificial neural networks for classification. Furthermore, achieving remarkable results is possible by employing specialized neural network configurations that incorporate the output of the BERT model and are tailored to the task. 

We not only utilize a variety of neural networks to classify the output of BERT-based models but can also leverage other SOTA models for classifying the contextualized language model's output. This provides us with a wide range of approaches, and we can subsequently distill the model to create a more compact version suitable for small and medium-scale requirements. If there are better language models, our proposed approach can perform better. Thus, CafeBERT \cite{do2024vlue} can be a good option to improve the model due to CafeBERT better than PhoBERT and XLM-R on various Vietnamese NLP tasks. Additionally, for an encoder-decoder architecture, we incorporate generative models like T5 \cite{t5}. To focus solely on decoder-related tasks, the GPT model \cite{gpt} proves to be the most suitable choice. Finally, BART \cite{bart} is highly effective, thanks to its popularity and promising potential for NLI tasks in the future. Furthermore, the ViNLI dataset is complete, consistent, reliable, challenging, and contextualized enough for pre-trained language models. This dataset is versatile and applicable to various NLP models, tasks, and sub-tasks.

\section*{Acknowledgement}
This research was supported by The VNUHCM-University of Information Technology's Scientific Research Support Fund.

\section*{Declarations}

\textbf{Conflict of interest} The authors declare that they have no conflict of interest.

\section*{Data Availability}

Data will be made available on reasonable request.

%\section*{Acknowledgments}
%We would like to express our deep gratitude to the University of Information Technology, Vietnam National University Ho Chi Minh City, for supporting us throughout this paper. We would also like to thank the editors and anonymous reviewers for their insightful comments and suggestions. Finally, the authors would like to thank Mr. Son T Luu and the 2019 VLSP Shared Task organizers for providing the datasets and assistance.

\bibliography{sn-bibliography}

\section{Appendix}
\textbf{ChatGPT prompt:} We prompt the input as the hypothesis-premise pair as follows.

"Tôi muốn bạn hóa thân thành chuyên gia suy luận ngữ ngôn tiếng việt. Tôi sẽ nhập vào 2 câu và bạn sẽ suy luận mối liên hệ giữa 2 câu đấy giữa các nhãn sau "sự kế thừa", "mâu thuẫn", "trung lập", "khác". Tôi muốn bạn chỉ trả lời nhãn suy luận mà không cần viết giải thích.\textbackslash n Câu 1: "\{hypothesis\}"\textbackslash n Câu 2: "\{premise\}"" ("I want you to act as a Vietnamese natural language inference expert. I will input 2 sentences, and you will infer the relationship between those sentences among the following labels: 'neutral', 'contradiction', 'neutral', 'other'. I want you to only provide the inferred labels without needing to write an explanation.\textbackslash n Sentence 1: "\{hypothesis\}"\textbackslash n Sentence 2: "\{premise\}"")

\end{document}